\documentclass[11pt,a4paper]{llncs}
\usepackage{graphicx}
\usepackage{url}

\begin{document}

\title{A Hypercat-enabled Semantic Internet of Things Data Hub: Technical Report}
\author{Ilias Tachmazidis\inst{1} \and Sotiris Batsakis\inst{1} \and John Davies\inst{2} \and Alistair Duke\inst{2} \and Mauro Vallati\inst{1} \and Grigoris Antoniou\inst{1}
\and Sandra Stincic Clarke\inst{2}
}

\institute{University of Huddersfield, Huddersfield, UK
\and
British Telecommunications, Ipswich, UK
}

\date{\today}
\maketitle

\begin{abstract}
An increasing amount of information is generated from the rapidly increasing number of sensor 
networks and smart devices. A wide variety of sources generate and publish information in
different formats, thus highlighting interoperability as one of the key prerequisites 
for the success of Internet of Things (IoT). The \textit{BT Hypercat Data Hub} provides a focal 
point for the sharing and consumption of available datasets from a wide range of sources. 
In this work, we propose a semantic enrichment of the \textit{BT Hypercat Data Hub}, using 
well-accepted Semantic Web standards and tools. We propose an ontology that captures 
the semantics of the imported data and present the \textit{BT SPARQL Endpoint} by means of a 
mapping between SPARQL and SQL queries. Furthermore, federated SPARQL queries allow queries over multiple   hub-based and external data sources. Finally, we provide two use cases in 
order to illustrate the advantages afforded by our semantic approach. 
\end{abstract}

\section{Introduction}\label{sec:Introduction}

The emerging notion of a smart city is based on the use of technology in order to 
improve the efficiency, effectiveness and capability of various city services, 
thus improving the quality of the inhabitants' lives~\cite{Townsend2013}. 
A fundamental difference between smart cities and similar uses 
of technology in other areas, such as business, government or education, is the vast 
variety of the technologies used, the types and volumes of data, and the services and 
applications targeted~\cite{Aquin2015}. Thus, developing successful smart city solutions
requires the collection and maintenance of relevant data in the form of IoT data.

Over the past few years, eight industry-led projects were funded by Innovate 
UK\footnote{\url{https://www.gov.uk/government/organisations/innovate-uk}} 
(the UK's innovation agency) to 
deliver IoT `clusters', each centred around a data hub to aggregate and expose data feeds 
from multiple sensor types.  The system that has come to be known as the \textit{BT Hypercat Data Hub} 
was part of the Internet of Things Ecosystem 
Demonstrator\footnote{\url{https://connect.innovateuk.org/web/internet-of-things-ecosystem-demonstrator/overview}} programme. 

Addressing interoperability by focusing on how interoperability could be achieved between 
data hubs in different domains was a major objective of the programme. Hence, Hypercat~\cite{Hypercat_3_0}
was developed, which is a standard for representing and exposing Internet of Things data 
hub catalogues~\cite{IoTWN} over web technologies, to improve data discoverability and 
interoperability. Recent work~\cite{Hypercat_JIST}, proposed a semantic enrichment for the 
core of the Hypercat specification, namely an RDF-based~\cite{rdfSemantic} equivalent for a JSON-based catalogue.
Other IoT / smart city projects include Barcelona\footnote{\url{http://ibarcelona.bcn.cat/en/smart-cities}}, MK:Smart\footnote{\url{http://www.mksmart.org}} which uses the \textit{BT Hypercat Data Hub} that is Hypercat-enabled but not semantically enriched, and the D-CAT\footnote{\url{https://www.w3.org/TR/vocab-dcat/}} catalogue approach from W3C.

The main objective of this work is to achieve the semantic enrichment~\cite{berners-lee} of the data in the 
\textit{BT Hypercat Data Hub} and to provide access to the enriched data through a SPARQL 
endpoint~\cite{prud2008sparql}. Furthermore, adding reasoning capabilities and the ability to combine external 
data sources using federated queries are important aspects of the implemented system.

The \textit{BT Hypercat Data Hub} provides a focal point for the sharing and consumption 
of available datasets from a wide range of sources. In order to enable rapid responses, data in the
\textit{BT Hypercat Data Hub} is stored in relational databases. In this work, sensor, 
event, and location databases, i.e., databases containing information about sensor readings, 
events and locations are used. In order to provide a semantically richer mechanism of accessing 
the available datasets, the \textit{BT Hypercat Ontology} was developed in order to lift semantically 
data stored within the relational databases. In addition, data translation through output 
adapters and SPARQL endpoints was defined. Thus, the semantically enriched data can be queried 
by accessing the developed \textit{BT SPARQL Endpoint}.

Triplestores contain the information in RDF format combined with a built-in SPARQL endpoint. Thus, triplestores are commonly used for providing SPARQL endpoints. However, as data 
in the \textit{BT Hypercat Data Hub} is stored in relational databases and this data is frequently 
updated, a more dynamic solution has been adopted. Thus, instead of copying the existing 
data into a triplestore, submitted SPARQL queries are dynamically translated into a set 
of SQL queries on top of the existing relational databases. In this way, a fully functioning 
SPARQL endpoint is provided, while during query execution, not only the SPARQL query itself 
is taken into consideration, but also the implicit information that is derived through reasoning 
over the developed ontology.

This work is organized as follows: Section~\ref{sec:Background} contains background information 
about the \textit{BT Hypercat Data Hub} prior to its semantic enrichment.
Section~\ref{sec:BT_Hypercat_Ontology} contains a 
description of the \textit{BT Hypercat Ontology} which was developted in this work in order to 
define the semantic representation of existing data. The corresponding mapping of data from a 
relational database to the semantic representation is described in Section~\ref{sec:Data_Translation}. 
The \textit{BT SPARQL Endpoint} is presented in Section~\ref{sec:BT_SPARQL_Endpoint} and the 
capability to combine information from external data sources by means of federated queries is 
presented in Section~\ref{sec:Federated_Querying}. Example use cases for the \textit{BT Hypercat Data Hub}  
are illustrated in Section~\ref{sec:Use_Cases}, while conclusions and future work are discussed 
in Section~\ref{sec:Conclusion}.

\section{Background}\label{sec:Background}

The role of the \textit{BT Hypercat Data Hub} is to enable information from a wide range of sources to be brought onto a common platform and presented to users and developers in a consistent way. Its portal provides a direct interface through which data consumers, such as app developers, can browse a data catalogue and select and subscribe to data feeds that they want to use. In addition, a JSON-based Hypercat~\cite{Hypercat_3_0} machine-readable catalogue, described further below, is also provided (as well as a recently proposed RDF-based Hypercat~\cite{Hypercat_JIST} catalogue). An API enables access to data feeds, secured by API keys, from browsers or within computer programs, while a relational, GIS capable, database enables complex queries that data can be filtered according to a wide range of criteria.  

A set of edge adapters enables information coming onto the hub to be converted to a standard format for use inside the platform's core.  It also provides a consistent API to end users and developers.  
The hub provides a consistent approach to integration between data exposed by sensors, systems and individuals via communication networks and the applications that can use derived information to improve decision making, e.g., in  control systems. It includes a set of adapters for ingress (input) and egress (output). These are potentially specific to each data source or application feed and may be implemented on a case by case basis. There is therefore a need to translate data between arbitrary external formats and the data formats used internally. 

In addition, as mentioned above, a Hypercat catalogue is implemented which is included via the Hypercat API. Hypercat is in essence a standard for representing and exposing Internet of Things data hub catalogues over web technologies, to improve data discoverability and interoperability. The idea is to enable distributed data repositories (data hubs) to be used jointly by applications through making it possible to query their catalogues in a uniform machine readable format. This enables the creation of “knowledge graphs” of available datasets across multiple hubs that applications can exploit and query to identify and access the data they need, whatever the data hub in which they are held. 

From this perspective, Hypercat represents a pragmatic starting point to solving the issues of managing multiple data sources, aggregated into multiple data hubs, through linked data and semantic web approaches. It incorporates a lightweight, JSON-based approach based on a technology stack used by a large population of web developers and as such offers a low barrier to entry. Hypercat allows a server (IoT hub) to provide a set of resources to a client, each with a set of metadata annotations. There are a small set of core mandatory metadata relations which a valid Hypercat catalogue must include; beyond this, implementers are free to use any set of annotations to suit their needs.

\section{BT Hypercat Ontology}\label{sec:BT_Hypercat_Ontology}

\begin{figure}[t]
  \centering
	\includegraphics[width=4.5in]{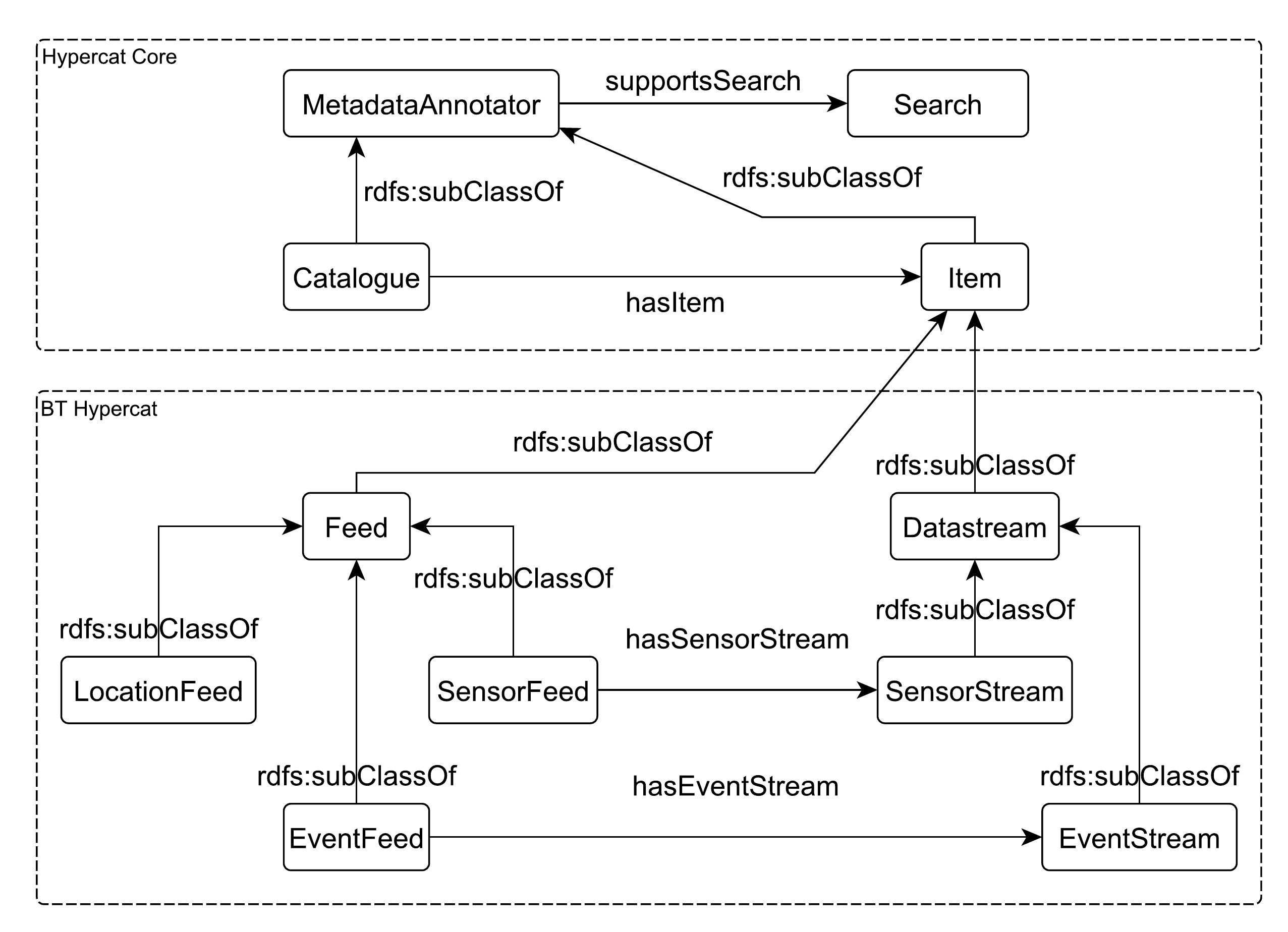}
  \caption{BT Hypercat Ontology.}
  \label{fig:BT_Hypercat_Ontology}
\end{figure}

In our previous work~\cite{Hypercat_JIST}, we proposed a semantic enrichment for the core
of the Hypercat specification, namely an RDF-based equivalent for a JSON-based catalogue.
While Hypercat offers a syntactic first step, providing semantically enriched data 
goes further by allowing the unique identification of existing resources, interoperability 
across various domains and further enrichment by combining internally stored data with the 
Linked Open Data (LOD) cloud\footnote{\url{http://lod-cloud.net/}}. Data enrichment in the 
\textit{BT Hypercat Data Hub} is achieved by representing data in RDF using concepts and 
properties defined in an OWL ontology \cite{OWL_Primer}. Figure~\ref{fig:BT_Hypercat_Ontology} shows the top 
level concepts of the \textit{BT Hypercat Ontology} and how the \textit{BT Hypercat Ontology}
extends the core of the Hypercat specification, following the proposed guidelines in~\cite{Hypercat_JIST}.

\textit{Feed} is the top level class for any data feed that is asserted in the knowledge base. It contains the semantic properties of feeds. These include the feed id, creator, update date, title, url, status, description, location name, domain and disposition. There are also subclasses of class \textit{Feed}, namely: \textit{SensorFeed}, \textit{EventFeed} and \textit{LocationFeed} representing feeds for sensors, events and locations respectively.

The modelled data has been incorporated in the \textit{BT Hypercat Data Hub} as one of the following feed types: (a) \textit{SensorFeed}, (b) \textit{EventFeed}, and (c) \textit{LocationFeed}. Practically, each data source can advertise available information through the \textit{BT Hypercat Data Hub} by providing a feed.  A feed should be understood as a source of sensor readings, events or locations. Within each feed, data is available through datastreams (a class \textit{Datastream} is defined, which has two subclasses namely: \textit{SensorStream} and \textit{EventStream} representing datastreams for sensors and events respectively). Thus, a given feed may provide a range of datastreams that are closely related e.g., for a weather data feed, different datastreams may provide sensor readings for temperature, humidity and visibility.
Considering information about locations, a feed (of type \textit{LocationFeed}) provides information directly by returning locations, namely locations are attached to and provided by a given feed. 

A Hypercat online catalogue\footnote{\url{http://portal.bt-hypercat.com/cat}}$^{,}$\footnote{\url{http://portal.bt-hypercat.com/cat-rdf}} contains details of feeds and information sources along with additional metadata such as tags, which allow improved search and discovery. The developed semantic model enables a semantic annotation and linkage of available feeds and datastreams. Thus, both classes \textit{Feed} and
\textit{Datastream} are modelled as subclasses of class \textit{Item} (which belongs to the core 
specification of Hypercat), namely feeds and datastreams are advertised as items of a given catalogue.
The \textit{BT Hypercat Ontology} has been developed and made available with the uri:
\begin{center}
	\begin{tabular}{ l }
        \url{http://portal.bt-hypercat.com/ontologies/bt-hypercat}
	\end{tabular}
\end{center}

\section{Data Translation}\label{sec:Data_Translation}

In this section we describe how data that is stored in a relational 
database within the \textit{BT Hypercat Data Hub}, is made available in RDF.

\subsection{RDF Adapter}\label{sec:RDF_Adapter}

By defining an ontology, semantically enriched data can be provided in RDF format. Note that prior to
the semantic enrichment only XML and JSON formats were available.
RDF data is represented in N-Triples format since such a format facilitates both storage and processing of data. Thus, each RDF triple is provided within a single line, in the following format: ``$<$subject$>$ $<$predicate$>$ $<$object$>$ .'', while a collection of RDF triples is stored as a collection of lines. Note that N-Triples format can easily be transformed into other valid RDF formats, such as RDF/XML. In addition, the generated knowledge base can also be loaded in any given triplestore, namely any given RDF knowledge base, in order to facilitate operations such as query answering. Thus, by following W3C standards  interoperability is ensured and the utilization of existing tools and applications is enabled. 

The \textit{BT Hypercat Data Hub} includes additional adapters for egress (output) in order to provide data in
RDF format. In the following, examples of how subject, predicate and object are generated for feeds and datastreams, are presented. Initially, the URI of each \textit{SensorFeed} is generated, namely:
\begin{center}
	\begin{tabular}{ l }
        $<$http://api.bt-hypercat.com/sensors/feeds/feedID$>$
	\end{tabular}
\end{center}
Note that ``http://api.bt-hypercat.com/'' is the prefix URI for any data provided by the \textit{BT Hypercat Data Hub}. 
In addition, ``/sensors'' provides information about the type of the feed (here 
\textit{SensorFeed}), followed by ``/feeds'', which indicates that this URI belongs to a resource 
describing a feed, and finally ``/feedID'' is an id that uniquely identifies the given feed. 
For each \textit{SensorFeed}, the \textit{BT Hypercat Data Hub} provides its type, namely:
\begin{center}
	\begin{tabular}{ | l | l |}
		\hline
		Subject & $<$http://api.bt-hypercat.com/sensors/feeds/feedID$>$ \\ \hline
		Predicate & $<$http://www.w3.org/1999/02/22-rdf-syntax-ns\#type$>$ \\ \hline
		Object & $<$http://portal.bt-hypercat.com/ontologies/\\
        & bt-hypercat\#SensorFeed$>$ \\
		\hline
	\end{tabular}
\end{center}

Each data property of \textit{SensorFeed} provides information in the following form 
(here is an example for property \textit{feed\_id}, other data properties are modelled in a 
similar fashion):
\begin{center}
	\begin{tabular}{ | l | l |}
		\hline
		Subject & $<$http://api.bt-hypercat.com/sensors/feeds/feedID$>$ \\ \hline
		Predicate & $<$http://portal.bt-hypercat.com/ontologies/\\
        & bt-hypercat\#feed\_id$>$ \\ \hline
		Object & ``feedID''$^{\wedge\wedge}$$<$http://www.w3.org/2001/\\
        & XMLSchema\#string$>$ \\
		\hline
	\end{tabular}
\end{center}

The URI of a given \textit{SensorStream} is generated as an extension of the URI of 
the \textit{SensorFeed} it belongs to, namely:
\begin{center}
	\begin{tabular}{ l }
        $<$http://api.bt-hypercat.com/sensors/feeds/feedID/datastreams/datastreamID$>$
	\end{tabular}
\end{center}
Here, ``/datastreams'' indicates that this URI belongs to a resource describing a datastream, 
and ``/datastreamID''  is the identifier of the given datastream. Thus, for each 
\textit{SensorStream}, the \textit{BT Hypercat Data Hub} provides its type, namely:
\begin{center}
	\begin{tabular}{ | l | l |}
		\hline
		Subject & $<$http://api.bt-hypercat.com/sensors/feeds/feedID/\\
        &datastreams/datastreamID$>$ \\ \hline
		Predicate & $<$http://www.w3.org/1999/02/22-rdf-syntax-ns\#type$>$ \\ \hline
		Object & $<$http://portal.bt-hypercat.com/ontologies/\\
        & bt-hypercat\#SensorStream$>$ \\
		\hline
	\end{tabular}
\end{center}

In addition, the fact that a given feed has a given datastream needs to be semantically annotated, 
namely the relation between \textit{SensorFeed} and \textit{SensorStream} is defined as follows:
\begin{center}
	\begin{tabular}{ | l | l |}
		\hline
		Subject & $<$http://api.bt-hypercat.com/sensors/feeds/feedID$>$ \\ \hline
		Predicate & $<$http://portal.bt-hypercat.com/ontologies/\\
        & bt-hypercat\#hasSensorStream$>$ \\ \hline
		Object & $<$http://api.bt-hypercat.com/sensors/feeds/feedID/\\
        & datastreams/datastreamID$>$ \\
		\hline
	\end{tabular}
\end{center}

In a similar way as for \textit{SensorFeed}, \textit{SensorStream} provides additional 
information through data properties (here is an example for property \textit{datastream\_id}, 
other data properties are modelled in a similar fashion):
\begin{center}
	\begin{tabular}{ | l | l |}
		\hline
		Subject & $<$http://api.bt-hypercat.com/sensors/feeds/feedID/ \\
        & datastreams/datastreamID$>$ \\ \hline
		Predicate & $<$http://portal.bt-hypercat.com/ontologies/\\
        & bt-hypercat\#datastream\_id$>$ \\ \hline
		Object & ``datastreamID''$^{\wedge\wedge}$$<$http://www.w3.org/2001/\\
        & XMLSchema\#string$>$ \\
		\hline
	\end{tabular}
\end{center}

A URI for \textit{EventFeed} is generated in a similar way as a URI for \textit{SensorFeed}, namely: 
\begin{center}
	\begin{tabular}{ l }
        $<$http://api.bt-hypercat.com/events/feeds/feedID$>$
	\end{tabular}
\end{center}
Note that the main difference is that ``/sensors'' is substituted by ``/events''. Thus, for each 
\textit{EventFeed}, the \textit{BT Hypercat Data Hub} provides its type, namely:
\begin{center}
	\begin{tabular}{ | l | l |}
		\hline
		Subject & $<$http://api.bt-hypercat.com/events/feeds/feedID$>$ \\ \hline
		Predicate & $<$http://www.w3.org/1999/02/22-rdf-syntax-ns\#type$>$ \\ \hline
		Object & $<$http://portal.bt-hypercat.com/ontologies/\\
        & bt-hypercat\#EventFeed$>$ \\
		\hline
	\end{tabular}
\end{center}

Each data property of \textit{EventFeed} provides information in the following form 
(here is an example for property \textit{feed\_id}, other data properties are modelled in a 
similar fashion):
\begin{center}
	\begin{tabular}{ | l | l |}
		\hline
		Subject & $<$http://api.bt-hypercat.com/events/feeds/feedID$>$ \\ \hline
		Predicate & $<$http://portal.bt-hypercat.com/ontologies/\\
        & bt-hypercat\#feed\_id$>$ \\ \hline
		Object & ``feedID''$^{\wedge\wedge}$$<$http://www.w3.org/2001/\\
        & XMLSchema\#string$>$ \\
		\hline
	\end{tabular}
\end{center}

A URI for \textit{EventStream} is generated in a similar way as a URI for \textit{SensorStream}.
The URI of a given \textit{EventStream} is generated as an extension of the URI of 
the \textit{EventFeed} it belongs to, namely:
\begin{center}
	\begin{tabular}{ l }
        $<$http://api.bt-hypercat.com/events/feeds/feedID/datastreams/datastreamID$>$
	\end{tabular}
\end{center}
Thus, for each \textit{EventStream}, the \textit{BT Hypercat Data Hub} provides its type, namely:
\begin{center}
	\begin{tabular}{ | l | l |}
		\hline
		Subject & $<$http://api.bt-hypercat.com/events/feeds/feedID/\\
        & datastreams/datastreamID$>$ \\ \hline
		Predicate & $<$http://www.w3.org/1999/02/22-rdf-syntax-ns\#type$>$ \\ \hline
		Object & $<$http://portal.bt-hypercat.com/ontologies/\\
        & bt-hypercat\#EventStream$>$ \\
		\hline
	\end{tabular}
\end{center}

In addition, the fact that a given feed has a given datastream needs to be semantically annotated, 
namely the relation between \textit{EventFeed} and \textit{EventStream} is defined as follows:
\begin{center}
	\begin{tabular}{ | l | l |}
		\hline
		Subject & $<$http://api.bt-hypercat.com/events/feeds/feedID$>$ \\ \hline
		Predicate & $<$http://portal.bt-hypercat.com/ontologies/\\
        & bt-hypercat\#hasEventStream$>$ \\ \hline
		Object & $<$http://api.bt-hypercat.com/events/feeds/feedID/\\
        & datastreams/datastreamID$>$ \\
		\hline
	\end{tabular}
\end{center}

In a similar way as for \textit{EventFeed}, \textit{EventStream} provides additional 
information through data properties (here is an example for property \textit{datastream\_id}, 
other data properties are modelled in a similar fashion):
\begin{center}
	\begin{tabular}{ | l | l |}
		\hline
		Subject & $<$http://api.bt-hypercat.com/events/feeds/feedID/\\
        & datastreams/datastreamID$>$ \\ \hline
		Predicate & $<$http://portal.bt-hypercat.com/ontologies/\\
        & bt-hypercat\#datastream\_id$>$ \\ \hline
		Object & ``datastreamID''$^{\wedge\wedge}$$<$http://www.w3.org/2001/\\
        & XMLSchema\#string$>$ \\
		\hline
	\end{tabular}
\end{center}

\subsection{SPARQL to SQL}\label{sec:SPARQL_to_SQL}

\begin{figure}[t]
  \centering
	\includegraphics[width=4.5in]{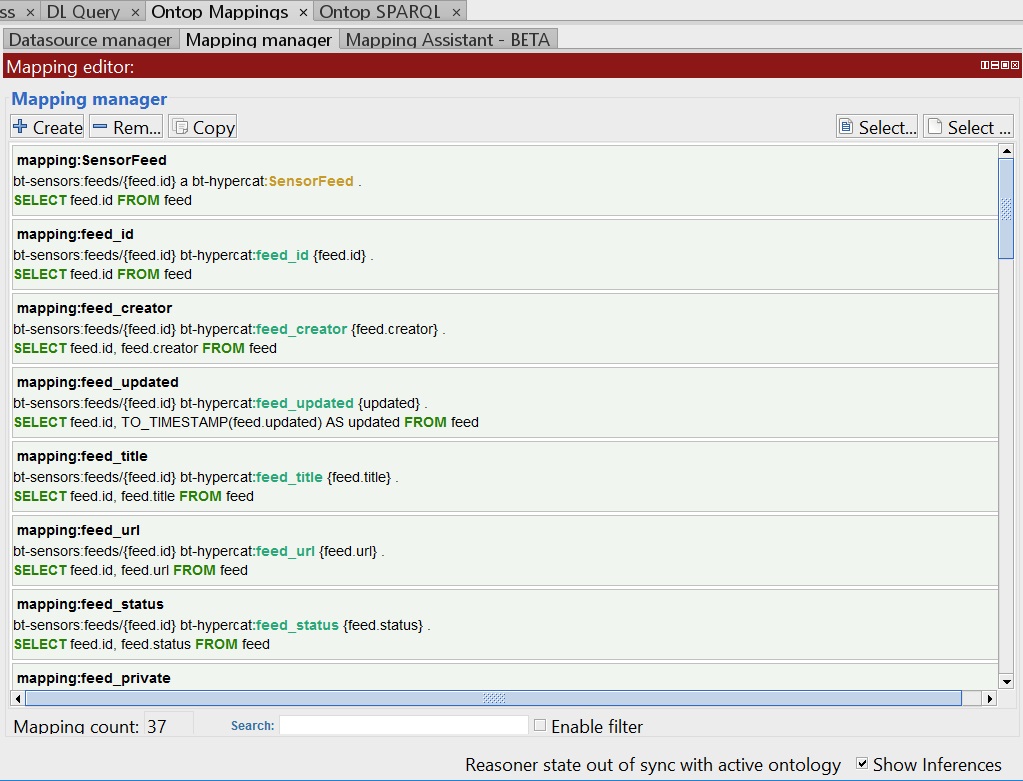}
  \caption{Protege Mapping Editor.}
  \label{fig:Protege_Mapping_Editor}
\end{figure}

In order to develop a \textit{SPARQL to SQL} endpoint,  Ontop\footnote{\url{http://ontop.inf.unibz.it/}}~\cite{Ontop} was used as an external library. Ontop comes with a Protege\footnote{\url{http://protege.stanford.edu/}} plug-in that allows the creation of mappings of SPARQL patterns to SQL queries (described below), see 
Figure~\ref{fig:Protege_Mapping_Editor}. 
In addition, it provides a reasoner that parses the mappings and the ontology, and handles the translation of SPARQL queries into a set of SQL queries in order to return the corresponding results (for the SPARQL query). A key advantage of using Ontop is that  implicit information that is extracted from the ontology through reasoning is taken into consideration. In this way,  semantically richer information compared to the knowledge that is stored in the relational database is provided. A description of how mappings can be created is presented below. 

In the following, an example of how a SPARQL triple pattern is mapped into a corresponding SQL query is described, and how the retrieved SQL results are used in order to construct RDF triples. \textit{Mapping ID} corresponds to a unique id for a given mapping, \textit{Target} (Triple Template) is the RDF triple pattern to be generated (note that SQL variables are given in braces, such as \{feed.id\}), and \textit{Source} (SQL Query) is the SQL query to be submitted to the database.

First,  the prefixes that are used are defined in order to shorten URIs, for example:
\begin{center}
	\begin{tabular}{ l }
		bt-sensors:	http://api.bt-hypercat.com/sensors/ \\
        bt-hypercat: http://portal.bt-hypercat.com/ontologies/bt-hypercat\#
	\end{tabular}
\end{center}
Then mappings are defined. 
For example, the following mapping maps the class \textit{SensorFeed}. Note that class \textit{SensorFeed} is subclass of \textit{Feed}, and thus is a valid assertion, while providing semantically richer information:

\begin{center}
	\begin{tabular}{ | l | l |}
		\hline
		Mapping ID & mapping:SensorFeed \\ \hline
		Target (Triple Template) & bt-sensors:feeds/\{feed.id\} \\ 
		& a \\ 
		& bt-hypercat:SensorFeed . \\ \hline
		Source (SQL Query) & SELECT feed.id \\
		& FROM feed \\
		\hline
	\end{tabular}
\end{center}
%Note that Figure~\ref{fig:Protege_Mapping_Editor} contains additional mappings for the 
%class \textit{SensorFeed}.

%The following mapping maps the data property \textit{feed\_id} of class \textit{Feed}:

%\begin{center}
%	\begin{tabular}{ | l | l |}
%		\hline
%		Mapping ID & mapping:feed\_id \\ \hline
%		Target (Triple Template) & bt-sensors:feeds/\{feed.id\} 
%bt-hypercat:feed\_id
%\{feed.id\} .\\ \hline
%		Source (SQL Query) & SELECT feed.id FROM feed \\
%		\hline
%	\end{tabular}
%\end{center}

The following query can be submitted to a \textit{SPARQL to SQL} endpoint in order 
to retrieve \textit{Feed}s:
\begin{verbatim}
PREFIX hypercat: <http://portal.bt-hypercat.com/ontologies/bt-hypercat#>
SELECT DISTINCT ?s
WHERE{ ?s a hypercat:Feed . }
\end{verbatim}

Thus, Ontop will match the triple pattern ``?s a hypercat:Feed'' with the mapping 
``mapping:SensorFeed'' since class \textit{SensorFeed} is subclass of \textit{Feed}.
An SQL query (see Source) will be submitted to the relational database, while the retrieved 
\textit{id}s (\textit{feed.id}) will be used in order to generate RDF triples following the triple
template (see Target).

%while the triple pattern ``?s bt-hypercat:feed\_id ?id'' will be matched with the
%mapping ``mapping:feed\_id''. Let us elaborate on mapping ``mapping:feed\_id'', 
%triple pattern ``?s bt-hypercat:feed\_id ?id'' matches ``mapping:feed\_id'' since
%they have a common predicate, namely ``bt-hypercat:feed\_id''.

The reader is referred to~\cite{Ontop} for more details on the internal functionality
of Ontop. Note that the generation of other triples follows a similar rational, while 
a detailed description of triple generation for a given concept or property can be found
in Appendix~\ref{sec:Appendix_mappings}.
%found in~\cite{ESWC_2017_Tech}.
%However, a detailed
%description of triple generation for each given concept and property is omitted due to
%space limitations, while further details of the can be found in~\cite{ESWC_2017_Tech}.

\section{BT SPARQL Endpoint}\label{sec:BT_SPARQL_Endpoint}
\begin{figure}[t]
  \centering
	\includegraphics[width=4.5in]{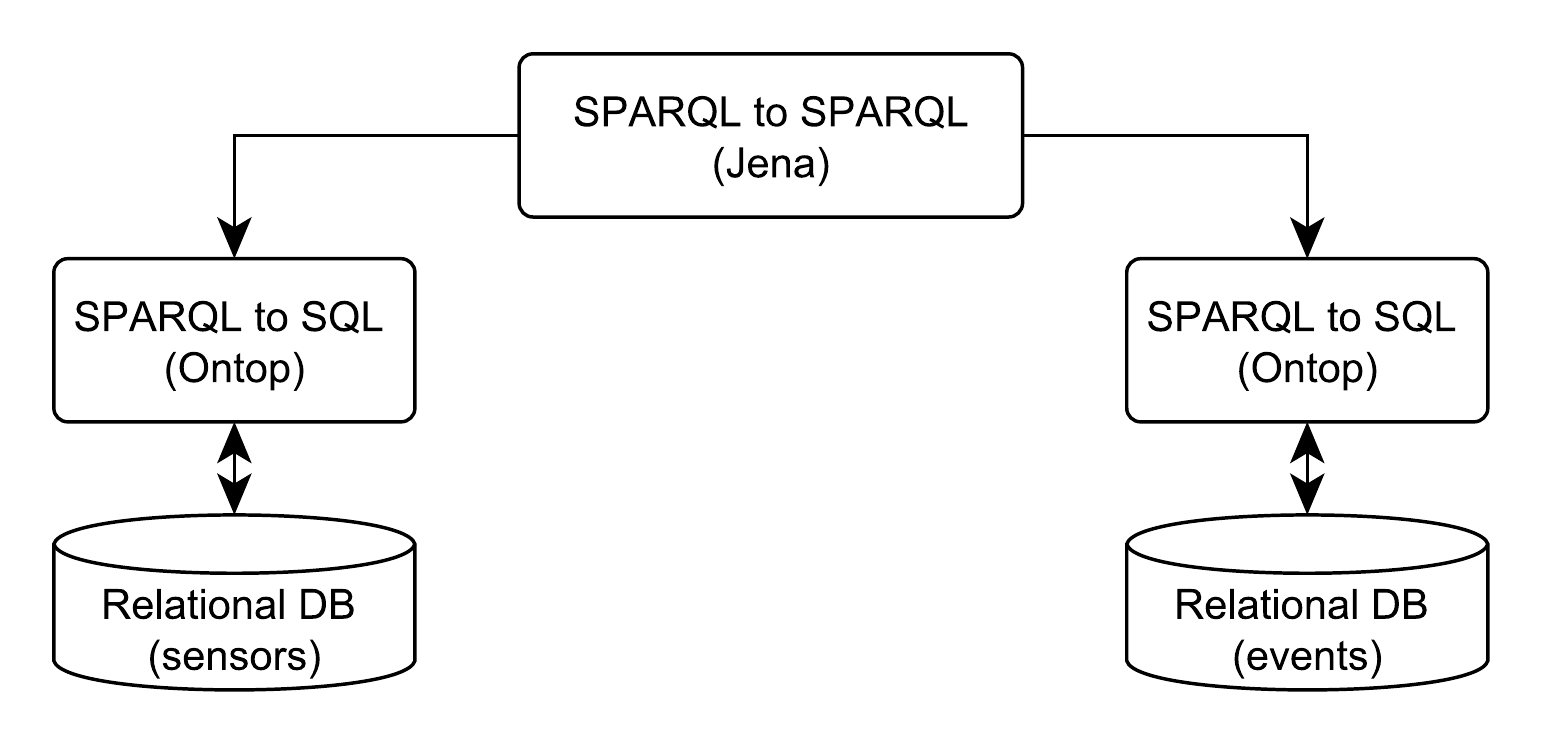}
  \caption{BT SPARQL Endpoint.}
  \label{fig:BT_SPARQL_Endpoint}
\end{figure}

In the following, a description of the high level architecture for the developed 
\textit{BT SPARQL Endpoint} is presented. As shown in Figure~\ref{fig:BT_SPARQL_Endpoint},  
two levels of abstraction are applied. At the lower level, there is a \textit{SPARQL to SQL} 
endpoint for each relational database in the system, namely each \textit{SPARQL to SQL} 
endpoint provides a SPARQL endpoint on top of the given relational database. In this way, 
the system administrator can add or remove a \textit{SPARQL to SQL} endpoint at any time. 

At the moment, a \textit{SPARQL to SQL} component is supporting the translation of SPARQL queries to PostgreSQL\footnote{\url{https://www.postgresql.org/}} relational databases that contain information about sensors or events. At the higher level, there is only one \textit{SPARQL to SPARQL} component (based on the query engine of Apache Jena\footnote{\url{https://jena.apache.org/index.html}}~\cite{carroll2004jena}), which is made available to end users. The underlying functionality indicates that end users submit SPARQL queries to the \textit{SPARQL to SPARQL} endpoint, while the system queries internally all available \textit{SPARQL to SQL} endpoints in order to extract the relevant information from existing relational databases. At any given point, the system administrator can add or remove a \textit{SPARQL to SQL} endpoint depending on the available PostgreSQL databases. 

\begin{figure}[t]
  \centering
	\includegraphics[width=4.5in]{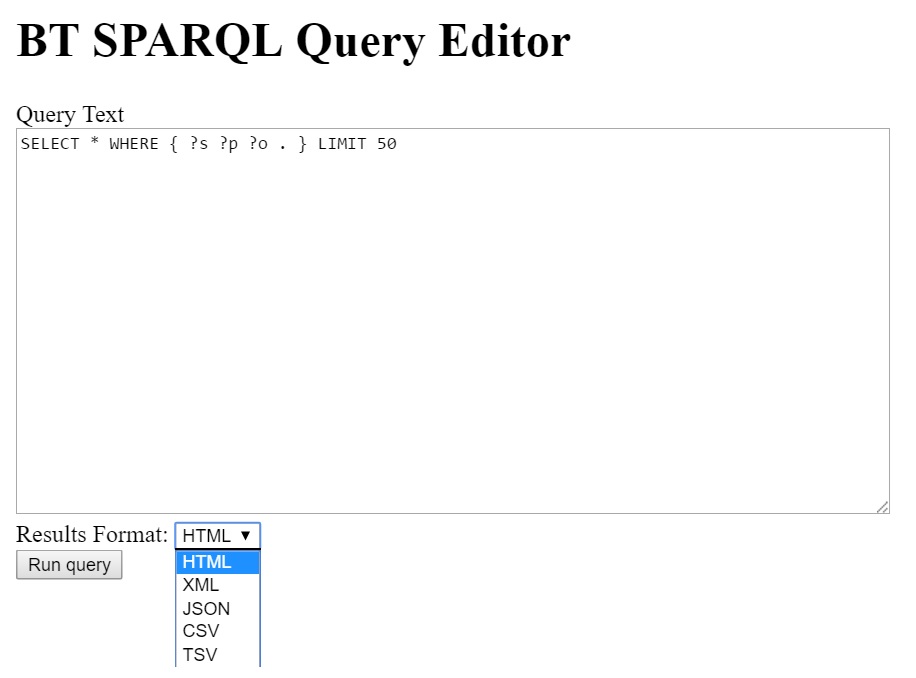}
  \caption{BT SPARQL Query Editor.}
  \label{fig:BT_SPARQL_Query_Editor}
\end{figure}

Both \textit{SPARQL to SPARQL} and \textit{SPARQL to SQL} endpoints can be accessed using the 
\textit{BT SPARQL Query Editor},  which is available for each endpoint. Users can provide the query 
text, namely the SPARQL query, as shown in Figure~\ref{fig:BT_SPARQL_Query_Editor}.
In addition, the \textit{BT SPARQL Query Editor} supports five results formats: HTML, XML, JSON, CSV and TSV. 

One of the key advantages of SPARQL queries over SQL queries is that SPARQL queries incorporate semantic reasoning within the returned results. For example, classes \textit{EventStream} and \textit{SensorStream} are subclasses of class \textit{Datastream}. Thus, the reasoner classifies all objects that belong to either \textit{EventStream} or \textit{SensorStream} as \textit{Datastream}. The following query can be submitted to a \textit{SPARQL to SPARQL} endpoint in order to retrieve  \textit{Datastream}s:
\begin{verbatim}
PREFIX hypercat: <http://portal.bt-hypercat.com/ontologies/bt-hypercat#>
SELECT DISTINCT ?s 
WHERE{ ?s a hypercat:Datastream . }
\end{verbatim}
Note that Ontop supports reasoning over RDFS\footnote{\url{http://www.w3.org/TR/rdf-schema/}} and 
OWL~2~QL\footnote{\url{https://www.w3.org/TR/owl-profiles/\#OWL\_2\_QL}}. 

\section{Federated Querying}\label{sec:Federated_Querying}

As described above, a \textit{Federated SPARQL} endpoint has been added in order to 
enable federated queries over both the \textit{BT SPARQL Endpoint} and other external SPARQL 
endpoints that are available through the LOD cloud. 
Such external SPARQL endpoints that are 
part of the LOD cloud are for example: DBPedia\footnote{\url{http://dbpedia.org/sparql}}, 
FactForge\footnote{\url{http://factforge.net/sparql}}, 
OpenUpLabs\footnote{\url{http://gov.tso.co.uk/transport/sparql}} and the European Environment 
Agency\footnote{\url{http://semantic.eea.europa.eu/sparql}}. 

\begin{figure}[t]
  \centering
	\includegraphics[width=4.5in]{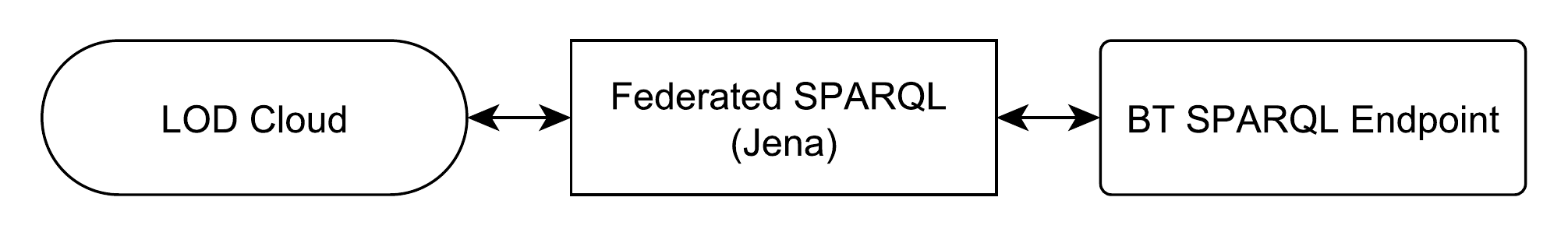}
  \caption{Federated SPARQL Endpoint.}
  \label{fig:Federated_SPARQL_Endpoint}
\end{figure}

The LOD cloud is expanding and new SPARQL endpoints are added (and removed) allowing for 
access to new data. Since the \textit{Federated SPARQL} endpoint does not contain any 
information itself, it serves as a middleware that combines information coming from other 
SPARQL endpoints, as depicted in Figure~\ref{fig:Federated_SPARQL_Endpoint}.

The \textit{Federated SPARQL} endpoint extends further the functionality of the \textit{BT SPARQL 
Endpoint} since external SPARQL endpoints can be used in order to retrieve information about 
events or social and economic information that can be combined with data from the \textit{BT SPARQL 
Endpoint} for complex data analytics. Examples can be the extraction of data about natural 
disasters from external datasets combined with related sensor and event 
data from the \textit{BT SPARQL Endpoint}. Other types of data extracted from external datasets 
can be, for example, social data related to housing projects and their correlation with sensor 
and event data from the \textit{BT SPARQL Endpoint}.

Reasoning capabilities and spatiotemporal queries can be combined with external datasets (LOD) 
in order to retrieve information which is not directly represented in the \textit{BT Hypercat Data Hub}. 
This can be achieved by means of federated queries spanning over different internal and external 
SPARQL endpoints.

For example, the following federated query retrieves sensor measurements from the \textit{BT Hypercat Data Hub} related to a specific active bus stop, extracted from an external SPARQL endpoint (OpenUpLabs): 
 
\begin{verbatim}
PREFIX geo: <http://www.w3.org/2003/01/geo/wgs84_pos#> 
PREFIX hypercat: <http://portal.bt-hypercat.com/ontologies/bt-hypercat#> 
PREFIX naptan: <http://transport.data.gov.uk/def/naptan/> 
PREFIX skos: <http://www.w3.org/2004/02/skos/core#>   

SELECT distinct ?d ?at_time ?western_longitude ?southern_latitude 
       ?eastern_longitude ?northern_latitude ?stop ?lat ?long 
WHERE { 
   SERVICE <http://gov.tso.co.uk/transport/sparql> 
   { 
      ?stop a naptan:CustomBusStop;    
            naptan:naptanCode ?naptanCode; 
            naptan:stopValidity ?stopValidity; 
            naptan:street "Kingswood Road"; 
            geo:lat ?lat; 
            geo:long ?long. 
      ?stopValidity naptan:stopStatus ?stopStatus. 
      ?stopStatus skos:prefLabel "Active"@en.  
   } 
   SERVICE <http://portal.bt-hypercat.com/BT-SPARQL-Endpoint/sparql> 
    { 
      ?d a hypercat:Datapoint. 
      ?d hypercat:datapoint_at_time ?at_time. 
      ?d hypercat:datapoint_western_longitude ?western_longitude.  
      ?d hypercat:datapoint_southern_latitude ?southern_latitude. 
      ?d hypercat:datapoint_eastern_longitude ?eastern_longitude. 
      ?d hypercat:datapoint_northern_latitude ?northern_latitude. 
      FILTER (?western_longitude > ?long - 0.1) 
      FILTER (?southern_latitude > ?lat - 0.1) 
      FILTER (?eastern_longitude < ?long + 0.1) 
      FILTER (?northern_latitude < ?lat + 0.1) 
   } 
   FILTER(BOUND(?d)) 
}
\end{verbatim}

The following federated query retrieves events from the 
\textit{BT Hypercat Data Hub} that took place close to an airport near 
London, extracted from an external SPARQL endpoint (FactForge): 
 
\begin{verbatim}
PREFIX geo: <http://www.w3.org/2003/01/geo/wgs84_pos#> 
PREFIX prop: <http://dbpedia.org/property/> 
PREFIX hypercat: <http://portal.bt-hypercat.com/ontologies/bt-hypercat#> 
PREFIX omgeo: <http://www.ontotext.com/owlim/geo#> 
PREFIX dbpediar: <http://dbpedia.org/resource/> 
PREFIX dbp-ont: <http://dbpedia.org/ontology/> 
PREFIX ff: <http://factforge.net/> 
PREFIX om: <http://www.ontotext.com/owlim/> 
 
SELECT distinct ?e ?event_date ?western_longitude ?southern_latitude 
       ?eastern_longitude ?northern_latitude ?label ?lat ?long 
WHERE { 
   SERVICE <http://factforge.net/sparql> 
   { 
      dbpediar:London geo:lat ?latBase; 
      geo:long ?longBase. 
      ?airport omgeo:nearby(?latBase ?longBase "50mi"); 
               a dbp-ont:Airport; 
               ff:preferredLabel ?label; 
               om:hasRDFRank ?RR; 
               geo:lat ?lat; 
               geo:long ?long.    
   } 
   SERVICE <http://portal.bt-hypercat.com/BT-SPARQL-Endpoint/sparql> 
   {   
      ?e a hypercat:Event. 
      ?e hypercat:event_sent ?event_date. 
      ?e hypercat:event_western_longitude ?western_longitude.  
      ?e hypercat:event_southern_latitude ?southern_latitude. 
      ?e hypercat:event_eastern_longitude ?eastern_longitude. 
      ?e hypercat:event_northern_latitude ?northern_latitude. 
      FILTER (?western_longitude > ?long - 0.5) 
      FILTER (?southern_latitude > ?lat - 0.5) 
      FILTER (?eastern_longitude < ?long + 0.5) 
      FILTER (?northern_latitude < ?lat + 0.5) 
   } 
   FILTER(BOUND(?e)) 
}  
\end{verbatim}

The following federated query retrieves events from the 
\textit{BT Hypercat Data Hub} that took place 
before a pollutant release, extracted from an external SPARQL 
endpoint (European Environment Agency): 
\begin{verbatim}
PREFIX hypercat: <http://portal.bt-hypercat.com/ontologies/bt-hypercat#> 
PREFIX xsd:   <http://www.w3.org/2001/XMLSchema#> 
PREFIX purl: <http://purl.org/dc/terms/> 
 
SELECT distinct ?e ?event_date ?western_longitude ?southern_latitude 
       ?eastern_longitude ?northern_latitude ?t ?date 
WHERE { 
   SERVICE <http://semantic.eea.europa.eu/sparql> 
   { 
      ?s purl:title ?t. 
      ?s purl:issued ?date 
      FILTER(regex(str(?t)," Pollutant "))   
   }  
   SERVICE <http://portal.bt-hypercat.com/BT-SPARQL-Endpoint/sparql>
   { 
      ?e a hypercat:Event. 
      ?e hypercat:event_sent ?event_date.  
      ?e hypercat:event_western_longitude ?western_longitude.  
      ?e hypercat:event_southern_latitude ?southern_latitude. 
      ?e hypercat:event_eastern_longitude ?eastern_longitude. 
      ?e hypercat:event_northern_latitude ?northern_latitude. 
      FILTER(BOUND(?e)) 
   }  
   FILTER(xsd:integer(year(xsd:dateTime(?date))) > 
          xsd:integer(year(xsd:dateTime(?event_date))))  
} 
\end{verbatim}

\section{Use Cases}\label{sec:Use_Cases}

This section is devoted to the description of two example use cases of the  \textit{BT Hypercat Data Hub}.

\subsection{The SimplifAI Project}

Urban traffic management and control is a primary concern of
any city, and urban traffic transport operators often have at their disposal a disparate
variety of real time and historical data, traffic controls (the most common of which are
traffic signals) and controlling software. Software systems used for traffic management have a vertical design: they are not integrated at a horizontal level and cannot therefore easily share their data, or exploit data provided from other software/sources.

For achieving a higher level of data integration, and to better capture and exploit real-time and historical urban data sources, the SimplifAI project was carried out by a consortium consisting of the University of Huddersfield, British Telecommunications, Transport for Greater Manchester, and two other SMEs. 
%The
%overall aims were in the context of developing smart city technology, taking advantage
%of the wide range of data available in a modern urban area. 
In particular, the project
focussed on exploiting the real-time and historical data sources to pursue better congestion control. As study area, a region of greater Manchester, UK was selected.

 %, in order to make data available from a common platform. The data made available for the SimplifAi project indicates the number of vehicles passing the site, e.g. a junction, in the previous

%5 minute period. A separate count for each road is provided as a datastream of the
%feed. The feeds are available via the data catalogue.

The overall concept in the improvement of traffic management was to utilise the semantically enriched data to enable the use of an
intelligent function which requires both the integration of traffic data from disparate
sources, and the transformation of the data into a predicate logic level, in order to
operate. The intelligent function was to create traffic signal strategies in real time to
solve challenges caused by exceptional or unexpected conditions.

The initial steps of the SimplifAI project concentrated on the semantic enrichment
of traffic data. The raw data was taken from a large number of transport and environment 
sources and integrated into the \textit{BT Hypercat Data Hub}, using the mapping of 
Section~\ref{sec:Data_Translation}. After that, the focus was put on the utilisation of 
semantic data for generating traffic control strategies. 

\begin{figure}[t]
\centering
\includegraphics[width=1\textwidth]{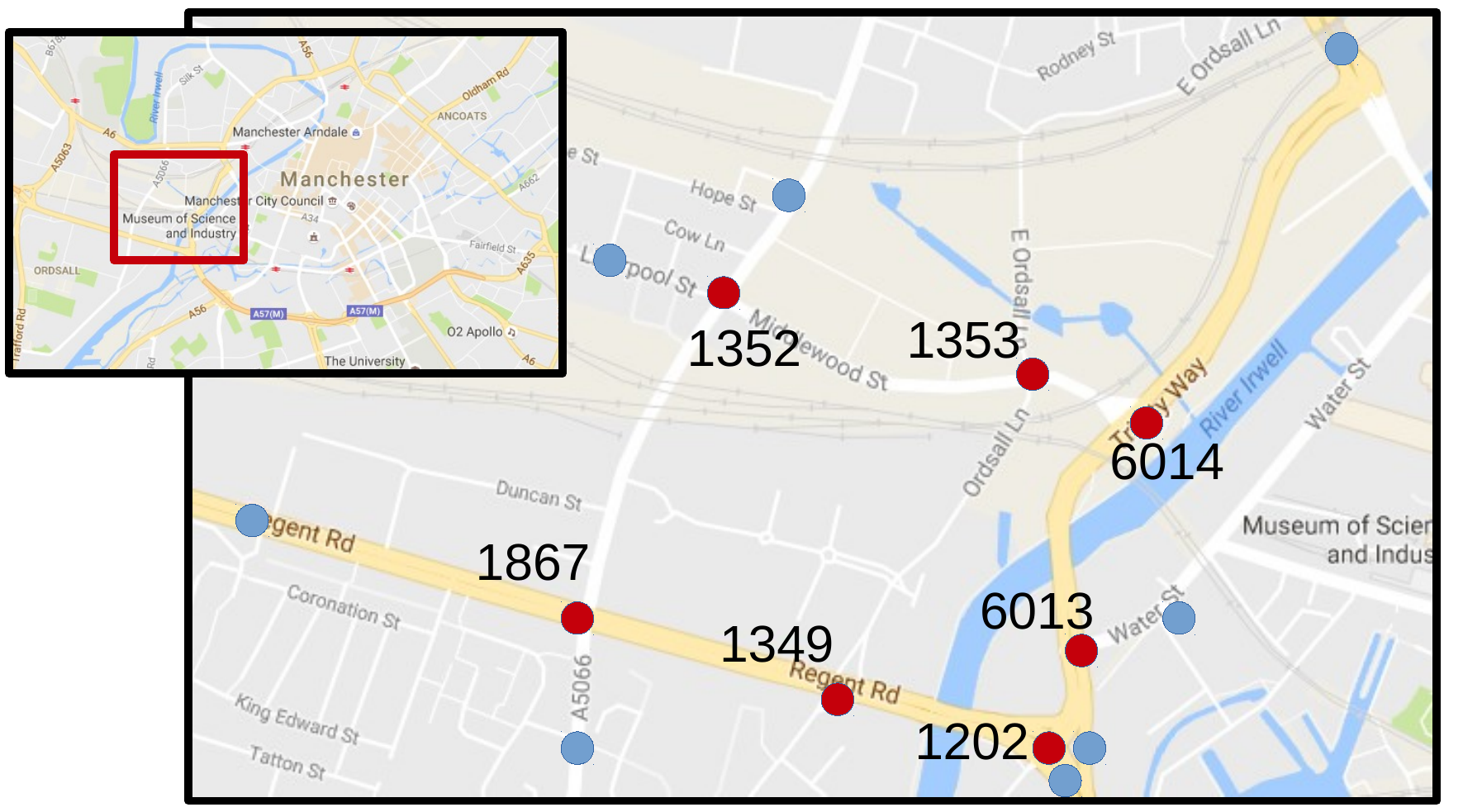}
\caption{The Modelled Area (large picture) and the position of the modelled area with regards to the Manchester city centre (small
picture, red-limited area). Red points indicate controlled junctions; blue points indicate the edge of the region.}\label{map1}
\end{figure}

By enriching semantically the imported data, the unique identification of imported data is enabled. 
This is orthogonal to the problem solved by planning, as planning can also deal with ad hoc data. 
However, once the study area expands, using semantically enriched data will allow a systematic 
way of identifying resources that are mentioned in the generated plans. In addition, federated 
queries
%(see Section~\ref{sec:Federated_Querying}) 
allow the developed system to extract data from the LOD cloud
%external data sources (i.e., LOD cloud) 
and combine it with data stored in the \textit{BT Hypercat Data Hub}
%(i.e., the \textit{BT Hypercat Data Hub}). 
%For example, 
(e.g., the federated query of Section~\ref{sec:Federated_Querying} combines bus stop 
information from an external source with internally stored data).
%is used for 
%combining  ).
%the federated query of

The intelligent function was based on an Automated Planning \cite{NauGhallabTraverso2004} approach \cite{aaai16}, that is able to generate traffic control strategies (actions which change signals
at a specified time) to alleviate traffic congestion caused by exceptional circumstances. The initial state of the modelled urban area, and information about available traffic lights and the structure of the network, were provided to the planning approach by the \textit{BT Hypercat Data Hub}. Figure \ref{map1} shows the map of the modelled area, in terms of junctions controlled by the planner (red points), links between junctions, and the boundaries of the area (blue dots). Boundaries are sources (destination) of incoming (outcoming) traffic flows. The planner was then executed in order to generate control strategies for a number of test scenarios, which were focussed on handling unexpected events. 

The quality of the strategies output from the planner was evaluated firstly by hand, inspecting the strategies to check that they were sensible, and by simulating their execution using traffic simulation software. Experts verified that strategies are sensible, and follow what would be expected when using ``common sense''.  Simulations confirmed that generated strategies can effectively deal with unexpected conditions better than standard urban traffic control approaches: on average, the area is de-congested 20\% faster, and tail-pipe emissions are reduced by 2.5\%. 

\subsection{City Concierge}

%\textbf{TODO: add use case description}

\begin{figure}[t]
  \begin{minipage}[t]{0.48\textwidth}
    \centering
    \includegraphics[width=2.0in]{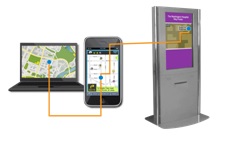}
    \caption{Interaction between end users and city wayfinding assets.}
    \label{fig:UI}
  \end{minipage}
  \hspace*{\fill} % it's important not to leave blank lines before and after this command
  \begin{minipage}[t]{0.48\textwidth}
    \centering
    \includegraphics[width=2.0in]{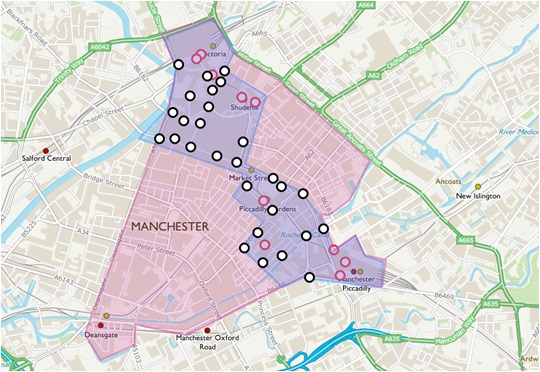}
    \caption{Locations of wayfinding infrastructure.}
    \label{fig:Potential_Trial_Area}
  \end{minipage}
\end{figure}

%CityVerve is a Manchester, UK based IoT Demonstrator project. It brings together the latest Internet of Things (IoT) technologies, deployed at city scale to deliver transformative benefits: new business and jobs for Manchester; better healthcare, transport and education; safer streets; and more engaged and empowered citizens. It was established in July 2016 with a two-year remit to demonstrate the capability of IoT applications and address barriers to deploying smart cities, such as city governance, network security, user trust and adoption, interoperability, scalability and justifying investment. Combining technology and business model innovations, it will create a real-life blueprint for smart cities worldwide.

%One of the use cases of the Cityverve project, City Concierge, is aiming to increase uptake of walking and cycling as a preferred travel mode in Greater Manchester. Currently, Greater Manchester lacks integrated, consistent wayfinding services that can be accessed through a variety of media, including digital and print. The City Concierge aims to develop a city user interface for the city region, integrating transportation and visitor services, allowing users to make informed choices regarding the way they travel. By ‘gluing’ together the end-to-end journeys, the City Concierge will harmonise the user experience across all modes, communications channels and media.

CityVerve is a Manchester, UK based IoT Demonstrator project, established in July 2016 
with a two-year focus on demonstrating the capability of IoT applications for smart cities.
One of the use cases of the CityVerve project, City Concierge, is aiming to increase uptake of walking and cycling as a preferred travel mode in Greater Manchester. Currently, Greater Manchester lacks integrated, consistent wayfinding services that can be accessed through a variety of media, including digital and print. 

The City Concierge aims to develop a city user interface for the city region, integrating transportation and visitor services, allowing users to make informed choices regarding the way they travel. %By ‘gluing’ together the end-to-end journeys, the City Concierge will harmonise the user experience across all modes, communications channels and media.
%Challenges for the use case are on many levels - technical, business and behavioural. 
The scope of the use case includes improvements in the way people navigate around the city with a digital solution in conjunction with physical wayfinding assets, see Figures~\ref{fig:UI} and~\ref{fig:Potential_Trial_Area}. %At the same time it will provide more tailored information to the travellers of the city, and allow the end user to make informed decisions.

%It is critical to understand the most efficient way to provide information to the public. Simultaneously, gaining insight and understanding on how people navigate around the city will be used to refine and improve the use case. Through IoT-enabled infrastructure, digital wayfinding infrastructure will be deployed that has the ability to sense pedestrians and provide updates on where to point, and what to display. CityVerve will support the wide GM-Connect wayfinding initiative. GM-Connect is a physical consistence wayfinding solution to be deployed in Greater Manchester, providing a seamless wayfinding link between Piccadilly station and Victoria station. Aligning with GM-Connect makes sense as the proposed approach reduces the maintenance costs associated with updating wayfinding information for a rapidly changing city environment. 

Currently, it has been established that the \textit{BT Hypercat Data Hub} provides the 
required infrastructure and functionality in order to enable the City Concierge. Translating 
data into RDF enables additional query capabilities such as SPARQL queries on top of the 
developed system and its combination with the LOD cloud through federated queries. Such 
queries are vital in order to achieve project's objectives, which include
%.Within the project, the objective is to 
the deployment of IoT and digital software solutions that seek to address current challenges, 
while having the flexibility for future solutions to be 
%delivered and 
developed on the network deployed as part of the CityVerve project.

The City Concierge use case could be enhanced by the use of SPARQL queries that allow data to be integrated into the application in a much more flexible way. One aim of the City Concierge is to inform travellers about cultural events that are happening in the city. This can be supported by a SPARQL query that queries for events and relies on an ontological reasoning to discover concerts, performances, art shows, exhibitions, etc. which can all be modelled as a subclass of \textit{Event} in a domain ontology. New types of event would be discovered without knowing in advance what type of event they are. Furthermore a federated SPARQL query could be used to discover local events that are described in a number of different SPARQL endpoints. %That will be achieved by running a number of demonstrators to show the capability of the open source network, and enabling SMEs and local people to deliver applications and solutions that fit their needs.

\section{Conclusion}\label{sec:Conclusion}
In this work, the semantic enrichment of the \textit{BT Hypercat Data Hub} has been presented. More
specifically, the \textit{BT Hypercat Ontology} has been introduced, which is the basis for the translation
of existing data into an RDF representation. In addition, the \textit{BT SPARQL Endpoint} has been 
implemented as a set of SPARQL endpoints and an additional endpoint, called \textit{Federated SPARQL}
endpoint, has been provided in order to allow the execution of federated queries. Moreover, several example federated queries illustrate how the \textit{BT Hypercat Data Hub} can be connected 
to the LOD cloud. Finally, two use cases are illustrating the extended functionality of the system,
thus highlighting the benefits of the semantic enrichment. 

Future work includes further semantic enrichment of the implemented system. Specifically, current
support for SPARQL queries can be extended in order to enable GeoSPARQL queries \cite{perry2012ogc} so as to provide 
direct access to spatial information that is currently available in the \textit{BT Hypercat Data Hub}.
In addition, spatiotemporal reasoning \cite{stravoskoufos2016sowl} is a prominent direction that could provide richer 
knowledge by reasoning over data that is coming from both the \textit{BT Hypercat Data Hub} and
the LOD cloud.

\bibliographystyle{plain}
\bibliography{main}

\appendix

\section{Appendix}\label{sec:Appendix_mappings}

Here we provide a detailed description of mappings for \textit{Feed}s
and \textit{Datastream}s. Note that in the developed system, mappings 
for sensors and events are implemented separately. The following prefixes 
are used in order to shorten URIs:
\begin{center}
	\begin{tabular}{ l }
		bt-sensors:	http://api.bt-hypercat.com/sensors/ \\
        bt-events:	http://api.bt-hypercat.com/events/ \\
        bt-hypercat: http://portal.bt-hypercat.com/ontologies/bt-hypercat\# \\
		wgs84\_pos:	http://www.w3.org/2003/01/geo/wgs84\_pos\#
	\end{tabular}
\end{center}

The following mapping maps the class \textit{SensorFeed}. Note that class \textit{SensorFeed} is subclass of \textit{Feed}, and thus is a valid assertion, while providing semantically richer information:

\begin{center}
	\begin{tabular}{ | l | l |}
		\hline
		Mapping ID & mapping:SensorFeed \\ \hline
		Target (Triple Template) & bt-sensors:feeds/\{feed.id\} \\
		& a \\ 
		& bt-hypercat:SensorFeed . \\ \hline
		Source (SQL Query) & SELECT feed.id FROM feed \\
		\hline
	\end{tabular}
\end{center}

The following mapping maps the class \textit{EventFeed}. Note that class \textit{EventFeed} is subclass of \textit{Feed}, and thus is a valid assertion, while providing semantically richer information:

\begin{center}
	\begin{tabular}{ | l | l |}
		\hline
		Mapping ID & mapping:EventFeed \\ \hline
		Target (Triple Template) & bt-events:feeds/\{feed.id\} \\
		& a \\
		& bt-hypercat:EventFeed . \\ \hline
		Source (SQL Query) & SELECT feed.id FROM feed \\
		\hline
	\end{tabular}
\end{center}

All data properties of both classes \textit{SensorFeed} and \textit{EventFeed} belong to 
their superclass, namely class \textit{Feed}. Thus, the mappings for class \textit{SensorFeed}
are provided, while the corresponding mappings for class \textit{EventFeed} can be defined
by substituting the prefix ``bt-sensors:'' with the prefix ``bt-events:''.
 
The following mapping maps the data property \textit{feed\_id} of class \textit{Feed}:

\begin{center}
	\begin{tabular}{ | l | l |}
		\hline
		Mapping ID & mapping:feed\_id \\ \hline
		Target (Triple Template) & bt-sensors:feeds/\{feed.id\} \\
		& bt-hypercat:feed\_id \\
		& \{feed.id\} .\\ \hline
		Source (SQL Query) & SELECT feed.id FROM feed \\
		\hline
	\end{tabular}
\end{center}

The following mapping maps the data property \textit{feed\_creator} of class \textit{Feed}:

\begin{center}
	\begin{tabular}{ | l | l |}
		\hline
		Mapping ID & mapping:feed\_creator \\ \hline
		Target (Triple Template) & bt-sensors:feeds/\{feed.id\} \\
		& bt-hypercat:feed\_creator \\
		& \{feed.creator\} .\\ \hline
		Source (SQL Query) & SELECT feed.id, feed.creator  \\
		& FROM feed \\
		\hline
	\end{tabular}
\end{center}

The following mapping maps the data property \textit{feed\_updated} of class \textit{Feed}:

\begin{center}
	\begin{tabular}{ | l | l |}
		\hline
		Mapping ID & mapping:feed\_updated \\ \hline
		Target (Triple Template) & bt-sensors:feeds/\{feed.id\} \\
		& bt-hypercat:feed\_updated \\
		& \{updated\} .\\ \hline
		Source (SQL Query) & SELECT feed.id, \\
		& TO\_TIMESTAMP(feed.updated) AS updated \\ 
		& FROM feed \\
		\hline
	\end{tabular}
\end{center}

The following mapping maps the data property \textit{feed\_title} of class \textit{Feed}:

\begin{center}
	\begin{tabular}{ | l | l |}
		\hline
		Mapping ID & mapping:feed\_title \\ \hline
		Target (Triple Template) & bt-sensors:feeds/\{feed.id\} \\
		& bt-hypercat:feed\_title \\
		& \{feed.title\} . \\ \hline
		Source (SQL Query) & SELECT feed.id, feed.title \\
		& FROM feed \\
		\hline
	\end{tabular}
\end{center}

The following mapping maps the data property \textit{feed\_url} of class \textit{Feed}:

\begin{center}
	\begin{tabular}{ | l | l |}
		\hline
		Mapping ID & mapping:feed\_url \\ \hline
		Target (Triple Template) & bt-sensors:feeds/\{feed.id\} \\
		& bt-hypercat:feed\_url \\
		& \{feed.url\} . \\ \hline
		Source (SQL Query) & SELECT feed.id, feed.url \\
		& FROM feed \\
		\hline
	\end{tabular}
\end{center}

The following mapping maps the data property \textit{feed\_status} of class \textit{Feed}:

\begin{center}
	\begin{tabular}{ | l | l |}
		\hline
		Mapping ID & mapping:feed\_status \\ \hline
		Target (Triple Template) & bt-sensors:feeds/\{feed.id\} \\
		& bt-hypercat:feed\_status \\
		& \{feed.status\} . \\ \hline
		Source (SQL Query) & SELECT feed.id, feed.status \\
		& FROM feed \\
		\hline
	\end{tabular}
\end{center}

The following mapping maps the data property \textit{feed\_private} of class \textit{Feed}:

\begin{center}
	\begin{tabular}{ | l | l |}
		\hline
		Mapping ID & mapping:feed\_private \\ \hline
		Target (Triple Template) & bt-sensors:feeds/\{feed.id\} \\
		& bt-hypercat:feed\_private \\
		& \{feed.private\} . \\ \hline
		Source (SQL Query) & SELECT feed.id, feed.private \\
		& FROM feed \\
		\hline
	\end{tabular}
\end{center}

The following mapping maps the data property \textit{feed\_description} of class \textit{Feed}:

\begin{center}
	\begin{tabular}{ | l | l |}
		\hline
		Mapping ID & mapping:feed\_description \\ \hline
		Target (Triple Template) & bt-sensors:feeds/\{feed.id\} \\
		& bt-hypercat:feed\_description \\
		& \{feed.description\} . \\ \hline
		Source (SQL Query) & SELECT feed.id, feed.description \\
		& FROM feed \\
		\hline
	\end{tabular}
\end{center}

The following mapping maps the data property \textit{feed\_icon} of class \textit{Feed}:

\begin{center}
	\begin{tabular}{ | l | l |}
		\hline
		Mapping ID & mapping:feed\_icon \\ \hline
		Target (Triple Template) &  bt-sensors:feeds/\{feed.id\} \\
		& bt-hypercat:feed\_icon \\
		& \{feed.icon\} . \\ \hline
		Source (SQL Query) & SELECT feed.id, feed.icon \\
		& FROM feed \\
		\hline
	\end{tabular}
\end{center}

The following mapping maps the data property \textit{feed\_website} of class \textit{Feed}:

\begin{center}
	\begin{tabular}{ | l | l |}
		\hline
		Mapping ID & mapping:feed\_website \\ \hline
		Target (Triple Template) & bt-sensors:feeds/\{feed.id\} \\
		& bt-hypercat:feed\_website \\
		& \{feed.website\} . \\ \hline
		Source (SQL Query) & SELECT feed.id, feed.website \\
		& FROM feed \\
		\hline
	\end{tabular}
\end{center}

The following mapping maps the data property \textit{feed\_email} of class \textit{Feed}:

\begin{center}
	\begin{tabular}{ | l | l |}
		\hline
		Mapping ID & mapping:feed\_email \\ \hline
		Target (Triple Template) & bt-sensors:feeds/\{feed.id\} \\
		& bt-hypercat:feed\_email \\
		& \{feed.email\} . \\ \hline
		Source (SQL Query) & SELECT feed.id, feed.email \\
		& FROM feed \\
		\hline
	\end{tabular}
\end{center}

The following mapping maps the data property \textit{feed\_tag} of class \textit{Feed}:

\begin{center}
	\begin{tabular}{ | l | l |}
		\hline
		Mapping ID & mapping:feed\_tag \\ \hline
		Target (Triple Template) & bt-sensors:feeds/\{feed.id\} \\
		& bt-hypercat:feed\_tag \\
		& \{tag\} . \\ \hline
		Source (SQL Query) & SELECT feed.id, \\
		& unnest(feed.tag) AS tag \\
		& FROM feed \\
		\hline
	\end{tabular}
\end{center}

The following mapping maps the data property \textit{feed\_location\_name} of class \textit{Feed}:

\begin{center}
	\begin{tabular}{ | l | l |}
		\hline
		Mapping ID & mapping:feed\_location\_name \\ \hline
		Target (Triple Template) & bt-sensors:feeds/\{feed.id\} \\
		& bt-hypercat:feed\_location\_name \\
		& \{feed.location\_name\} . \\ \hline
		Source (SQL Query) & SELECT feed.id, feed.location\_name \\
		& FROM feed \\
		\hline
	\end{tabular}
\end{center}

The following mapping maps the data property \textit{feed\_exposure} of class \textit{Feed}:

\begin{center}
	\begin{tabular}{ | l | l |}
		\hline
		Mapping ID & mapping:feed\_exposure \\ \hline
		Target (Triple Template) & bt-sensors:feeds/\{feed.id\} \\
		& bt-hypercat:feed\_exposure \\
		& \{feed.exposure\} . \\ \hline
		Source (SQL Query) & SELECT feed.id, feed.exposure \\
		& FROM feed \\
		\hline
	\end{tabular}
\end{center}

The following mapping maps the data property \textit{feed\_domain} of class \textit{Feed}:

\begin{center}
	\begin{tabular}{ | l | l |}
		\hline
		Mapping ID & mapping:feed\_domain \\ \hline
		Target (Triple Template) & bt-sensors:feeds/\{feed.id\} \\
		& bt-hypercat:feed\_domain \\
		& \{feed.dom\} . \\ \hline
		Source (SQL Query) & SELECT feed.id, feed.dom \\
		& FROM feed \\
		\hline
	\end{tabular}
\end{center}

The following mapping maps the data property \textit{feed\_disposition} of class \textit{Feed}:

\begin{center}
	\begin{tabular}{ | l | l |}
		\hline
		Mapping ID & mapping:feed\_disposition \\ \hline
		Target (Triple Template) & bt-sensors:feeds/\{feed.id\} \\
		& bt-hypercat:feed\_disposition \\
		& \{feed.disposition\} . \\ \hline
		Source (SQL Query) & SELECT feed.id, feed.disposition \\
		& FROM feed \\
		\hline
	\end{tabular}
\end{center}

The following mapping maps the data property \textit{feed\_lat} of class \textit{Feed}
as \textit{wgs84\_pos:lat}:

\begin{center}
	\begin{tabular}{ | l | l |}
		\hline
		Mapping ID & mapping:feed\_lat \\ \hline
		Target (Triple Template) & bt-sensors:feeds/\{feed.id\} \\
		& wgs84\_pos:lat \\
		& \{feed.lat\} . \\ \hline
		Source (SQL Query) & SELECT feed.id, feed.lat \\
		& FROM feed \\
		\hline
	\end{tabular}
\end{center}

The following mapping maps the data property \textit{feed\_lon} of class \textit{Feed}
as \textit{wgs84\_pos:long}:

\begin{center}
	\begin{tabular}{ | l | l |}
		\hline
		Mapping ID & mapping:feed\_lon \\ \hline
		Target (Triple Template) & bt-sensors:feeds/\{feed.id\} \\
		& wgs84\_pos:long \\
		& \{feed.lon\} . \\ \hline
		Source (SQL Query) & SELECT feed.id, feed.lon \\
		& FROM feed \\
		\hline
	\end{tabular}
\end{center}

The following mapping maps the data property \textit{feed\_ele} of class \textit{Feed}
as \textit{wgs84\_pos:alt}:

\begin{center}
	\begin{tabular}{ | l | l |}
		\hline
		Mapping ID & mapping:feed\_ele \\ \hline
		Target (Triple Template) & bt-sensors:feeds/\{feed.id\} \\
		& wgs84\_pos:alt \\
		& \{feed.ele\} . \\ \hline
		Source (SQL Query) & SELECT feed.id, feed.ele \\ 
		& FROM feed \\
		\hline
	\end{tabular}
\end{center}

The following mapping maps the data property \textit{feed\_the\_geom} of class \textit{Feed}:

\begin{center}
	\begin{tabular}{ | l | l |}
		\hline
		Mapping ID & mapping:feed\_the\_geom \\ \hline
		Target (Triple Template) & bt-sensors:feeds/\{feed.id\} \\
		& bt-hypercat:feed\_the\_geom \\
		& \{the\_geom\} . \\ \hline
		Source (SQL Query) & SELECT feed.id, \\
		& ST\_AsText(feed.the\_geom) AS the\_geom \\
		& FROM feed \\
		\hline
	\end{tabular}
\end{center}

The following mapping maps the object property \textit{hasSensorStream} of class \textit{SensorFeed}:

\begin{center}
	\begin{tabular}{ | l | l |}
		\hline
		Mapping ID & mapping:hasSensorStream \\ \hline
		Target (Triple Template) & bt-sensors:feeds/\{datastream.feed\} \\
		& bt-hypercat:hasSensorStream \\
		& bt-sensors:feeds/\{datastream.feed\}/datastreams/\{datastream.id\} . \\ \hline
		Source (SQL Query) & SELECT datastream.feed, datastream.id \\
		& FROM datastream \\
		\hline
	\end{tabular}
\end{center}

The following mapping maps the data property \textit{hasEventStream} of class \textit{EventFeed}:

\begin{center}
	\begin{tabular}{ | l | l |}
		\hline
		Mapping ID & mapping:hasEventStream \\ \hline
		Target (Triple Template) & bt-events:feeds/\{datastream.feed\} \\
		& bt-hypercat:hasEventStream \\
		& bt-events:feeds/\{datastream.feed\}/datastreams/\{datastream.id\} . \\ \hline
		Source (SQL Query) & SELECT datastream.feed, datastream.id \\
		& FROM datastream \\
		\hline
	\end{tabular}
\end{center}

The following mapping maps the class \textit{SensorStream}. Note that class \textit{SensorStream} 
is subclass of \textit{Datastream}, and thus is a valid assertion, while providing semantically 
richer information:

\begin{center}
	\begin{tabular}{ | l | l |}
		\hline
		Mapping ID & mapping:SensorStream \\ \hline
		Target (Triple Template) & bt-sensors:feeds/\{datastream.feed\}/datastreams/\{datastream.id\} \\
		& a \\
		& bt-hypercat:SensorStream . \\ \hline
		Source (SQL Query) & SELECT datastream.feed, datastream.id \\
		& FROM datastream \\
		\hline
	\end{tabular}
\end{center}

The following mapping maps the class \textit{EventStream}. Note that class \textit{EventStream} 
is subclass of \textit{Datastream}, and thus is a valid assertion, while providing semantically 
richer information:

\begin{center}
	\begin{tabular}{ | l | l |}
		\hline
		Mapping ID & mapping:EventStream \\ \hline
		Target (Triple Template) & bt-events:feeds/\{datastream.feed\}/datastreams/\{datastream.id\} \\
		& a \\
		& bt-hypercat:EventStream . \\ \hline
		Source (SQL Query) & SELECT datastream.feed, datastream.id \\
		& FROM datastream \\
		\hline
	\end{tabular}
\end{center}

All data properties of class \textit{EventStream} are contained in class \textit{SensorStream} as
well. Thus, these data properties belong to their superclass, namely class \textit{Datastream}. 
However, class \textit{SensorStream} contains additional data properties that do not belong to 
class \textit{EventStream}. 

For data properties \textit{datastream\_id}, \textit{datastream\_tag},
\textit{datastream\_current\_time} and \textit{datastream\_current\_value}, the mappings for class \textit{SensorFeed} are provided, while the corresponding mappings for class \textit{EventFeed} 
can be defined by substituting the prefix ``bt-sensors:'' with the prefix ``bt-events:''. On the
other hand, for data properties \textit{datastream\_max\_value}, \textit{datastream\_min\_value},
\textit{datastream\_unit\_symbol}, \textit{datastream\_unit\_type} and \textit{datastream\_unit\_text}
the mappings are provided only for class \textit{SensorFeed} (these data properties do not belong
to class \textit{EventStream}).

The following mapping maps the data property \textit{datastream\_id} of class \textit{Datastream}:

\begin{center}
	\begin{tabular}{ | l | l |}
		\hline
		Mapping ID & mapping:datastream\_id \\ \hline
		Target (Triple Template) & bt-sensors:feeds/\{datastream.feed\}/datastreams/\{datastream.id\} \\
		& bt-hypercat:datastream\_id \\
		& \{datastream.id\} . \\ \hline
		Source (SQL Query) & SELECT datastream.feed, datastream.id \\
		& FROM datastream \\
		\hline
	\end{tabular}
\end{center}

The following mapping maps the data property \textit{datastream\_tag} of class \textit{Datastream}:

\begin{center}
	\begin{tabular}{ | l | l |}
		\hline
		Mapping ID & mapping:datastream\_tag \\ \hline
		Target (Triple Template) & bt-sensors:feeds/\{datastream.feed\}/datastreams/\{datastream.id\} \\
		& bt-hypercat:datastream\_tag \\
		& \{tag\} . \\ \hline
		Source (SQL Query) & SELECT datastream.feed, datastream.id, \\
		& unnest(datastream.tag) AS tag \\
		& FROM datastream \\
		\hline
	\end{tabular}
\end{center}

The following mapping maps the data property \textit{datastream\_current\_time} of class \textit{Datastream}:

\begin{center}
	\begin{tabular}{ | l | l |}
		\hline
		Mapping ID & mapping:datastream\_current\_time \\ \hline
		Target (Triple Template) & bt-sensors:feeds/\{datastream.feed\}/datastreams/\{datastream.id\} \\
		& bt-hypercat:datastream\_current\_time \\
		& \{current\_time\} . \\ \hline
		Source (SQL Query) & SELECT datastream.feed, datastream.id, \\
		& TO\_TIMESTAMP(datastream.c\_time) AS current\_time \\
		& FROM datastream \\
		\hline
	\end{tabular}
\end{center}

The following mapping maps the data property \textit{datastream\_current\_value} of class \textit{Datastream}:

\begin{center}
	\begin{tabular}{ | l | l |}
		\hline
		Mapping ID & mapping:datastream\_current\_value \\ \hline
		Target (Triple Template) & bt-sensors:feeds/\{datastream.feed\}/datastreams/\{datastream.id\} \\
		& bt-hypercat:datastream\_current\_value \\
		& \{datastream.c\_value\} .\\ \hline
		Source (SQL Query) & SELECT datastream.feed, datastream.id, datastream.c\_value \\
		& FROM datastream \\
		\hline
	\end{tabular}
\end{center}

The following mapping maps the data property \textit{datastream\_max\_value} of class \textit{SensorStream}:

\begin{center}
	\begin{tabular}{ | l | l |}
		\hline
		Mapping ID & mapping:datastream\_max\_value \\ \hline
		Target (Triple Template) & bt-sensors:feeds/\{datastream.feed\}/datastreams/\{datastream.id\} \\
		& bt-hypercat:datastream\_max\_value \\
		& \{datastream.max\_value\} . \\ \hline
		Source (SQL Query) & SELECT datastream.feed, datastream.id, datastream.max\_value \\
		& FROM datastream \\
		\hline
	\end{tabular}
\end{center}

The following mapping maps the data property \textit{datastream\_min\_value} of class \textit{SensorStream}:

\begin{center}
	\begin{tabular}{ | l | l |}
		\hline
		Mapping ID & mapping:datastream\_min\_value \\ \hline
		Target (Triple Template) & bt-sensors:feeds/\{datastream.feed\}/datastreams/\{datastream.id\} \\
		& bt-hypercat:datastream\_min\_value \\
		& \{datastream.min\_value\} . \\ \hline
		Source (SQL Query) & SELECT datastream.feed, datastream.id, datastream.min\_value \\
		& FROM datastream \\
		\hline
	\end{tabular}
\end{center}

The following mapping maps the data property \textit{datastream\_unit\_symbol} of class \textit{SensorStream}:

\begin{center}
	\begin{tabular}{ | l | l |}
		\hline
		Mapping ID & mapping:datastream\_unit\_symbol \\ \hline
		Target (Triple Template) & bt-sensors:feeds/\{datastream.feed\}/datastreams/\{datastream.id\} \\
		& bt-hypercat:datastream\_unit\_symbol \\
		& \{datastream.unit\_symbol\} . \\ \hline
		Source (SQL Query) & SELECT datastream.feed, datastream.id, datastream.unit\_symbol \\
		& FROM datastream \\
		\hline
	\end{tabular}
\end{center}

The following mapping maps the data property \textit{datastream\_unit\_type} of class \textit{SensorStream}:

\begin{center}
	\begin{tabular}{ | l | l |}
		\hline
		Mapping ID & mapping:datastream\_unit\_type \\ \hline
		Target (Triple Template) & bt-sensors:feeds/\{datastream.feed\}/datastreams/\{datastream.id\} \\ 
		& bt-hypercat:datastream\_unit\_type \\
		& \{datastream.unit\_type\} . \\ \hline
		Source (SQL Query) & SELECT datastream.feed, datastream.id, datastream.unit\_type \\ 
		& FROM datastream \\
		\hline
	\end{tabular}
\end{center}

The following mapping maps the data property \textit{datastream\_unit\_text} of class \textit{SensorStream}:

\begin{center}
	\begin{tabular}{ | l | l |}
		\hline
		Mapping ID & mapping:datastream\_unit\_text \\ \hline
		Target (Triple Template) & bt-sensors:feeds/\{datastream.feed\}/datastreams/\{datastream.id\} \\ 
		& bt-hypercat:datastream\_unit\_text \\
		& \{datastream.unit\_text\} . \\ \hline
		Source (SQL Query) & SELECT datastream.feed, datastream.id, datastream.unit\_text \\
		& FROM datastream \\
		\hline
	\end{tabular}
\end{center}

\end{document}